\documentclass[pdflatex,sn-mathphys-ay]{sn-jnl}
\usepackage{graphicx}%
\usepackage{multirow}%
\usepackage{amsmath,amssymb,amsfonts}%
\usepackage{amsthm}%
\usepackage{mathrsfs}%
\usepackage[title]{appendix}%
\usepackage{xcolor}%
\usepackage{textcomp}%
\usepackage{manyfoot}%
\usepackage{booktabs}%
\usepackage{algorithm}%
\usepackage{algorithmicx}%
\usepackage{algpseudocode}%
\usepackage{listings}%

\begin{document}

\title[Article Title]{Ancient but Digitized: Developing Handwritten Optical Character Recognition for East Syriac
Script Through Creating KHAMIS Dataset}

\author{\fnm{Ameer} \sur{Majeed\textsuperscript{1}}}\email{ameer.ha.majeed@gmail.com\textsuperscript{1}}

\author{\fnm{Hossein} \sur{Hassani\textsuperscript{2}}}\email{hosseinh@ukh.edu.krd\textsuperscript{2}}

\affil{\orgdiv{Department of Computer Science}, \orgname{University of Kurdistan Hewlêr}, \orgaddress{\city{Erbil}, \state{Kurdistan Region}, \country{Iraq}}}

\abstract{Many languages have vast amounts of handwritten texts, such as ancient scripts about folktale stories and historical
narratives or contemporary documents and letters. Digitization of those texts has various applications, such as daily tasks,
cultural studies, and historical research. Syriac is an ancient, endangered, and low-resourced language that has not
received the attention it requires and deserves. This paper reports on a research project aimed at developing a optical character recognition (OCR) model based on the handwritten Syriac texts as a starting point to build more digital services for this endangered language. A dataset was created, KHAMIS (inspired by the East Syriac poet, Khamis bar Qardahe), which consists of handwritten sentences in the East Syriac script. We used it to fine-tune the Tesseract-OCR engine's pretrained Syriac model on handwritten data. The data was collected from volunteers capable of reading and writing in the language to create KHAMIS. KHAMIS currently consists of 624 handwritten Syriac sentences collected from 31 university students and one professor, and it will be partially available online and the whole dataset available in the near future for development and research purposes. As a result, the handwritten OCR model
was able to achieve a character error rate of 1.097-1.610\% and 8.963-10.490\% on both training and
evaluation sets, respectively, and both a character error rate of 18.89-19.71\%
and a word error rate of 62.83-65.42\%  when evaluated on the test set, which is twice as better than the default Syriac
model of Tesseract.}


\maketitle

\section{Introduction}\label{sec1}

Optical character recognition (OCR) is a computational technique that is used to recognize text from scanned documents and digital images. As contemporary human society is undergoing a rapid growth and everyone predominantly relies on digital tools and technological resources to perform most of their daily tasks, there is a need for digitizing and shifting towards automated recognition of typed and handwritten documents. For instance, researchers from the field of digital humanities are in need of automated methods like OCR to search and analyze a large corpora in order to save time \citep{chesley}. This paper seeks to fill the gap in research related to handwritten Syriac OCR and to create an OCR model that can exceed an accuracy of 50\% trained on a custom handwritten dataset of different sentences collected by volunteers who can read and write in the language.

The paper is organized as follows: Section 2 will provide a brief overview of the Syriac language and its script, its importance to history, and why its preservation is of utmost importance. The third Section mainly discusses the previous works done in Syriac OCR and to pinpoint the gaps in literature, and it also
provides an overview of the research and the methods that have been used in Arabic, Persian, and other cursive scripts, be it a holistic approach or analytical. Based
on the conclusion of the preceding Section, the fourth Section concerns itself with
methodology and the step-by-step procedure that the OCR is going to go through, i.e, it will be divided into 5 subsections: data collection, preprocessing, a description on the proposed method, training the model, and the method of evaluation. The dataset is the topic of the fifth Section where a detailed process will be mentioned as to how it has been collected and what significance this dataset has. The sixth Section will be dedicated to the experiments that are going to be conducted on the image data in hand, and it will shed light on the results of the experiments and the performance of the OCR model on Syriac handwritten data. The seventh Section will be the discussion, of which will elaborate on the results and give an analysis on both the positives and the shortcomings of the OCR model. Lastly, the eighth Section is a summary of the work that has been done, and it will propose recommendations and insights so as to develop better handwritten OCR models for Syriac in the future.

\section{Syriac Language and Script - An Overview}
\begin{figure}[H]

        \centerline{\includegraphics[width=80mm,scale=1]{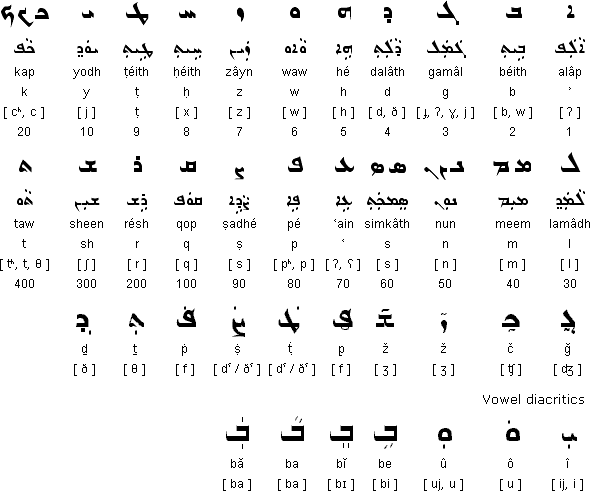}}
        \caption{East Syriac (Madnḥāyā) Script \citep{omniglot}}
        \label{fig}
        
    \end{figure}

\begin{figure}[H]
        \centerline{\includegraphics[width=80mm,scale=1]{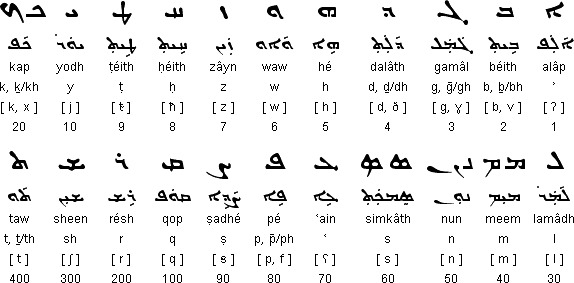}}
        \caption{Estrangela Script \citep{omniglot}}
        \label{fig2}

    \end{figure}

\begin{figure}[H]
    \centerline{\includegraphics[width=80mm,scale=1]
    {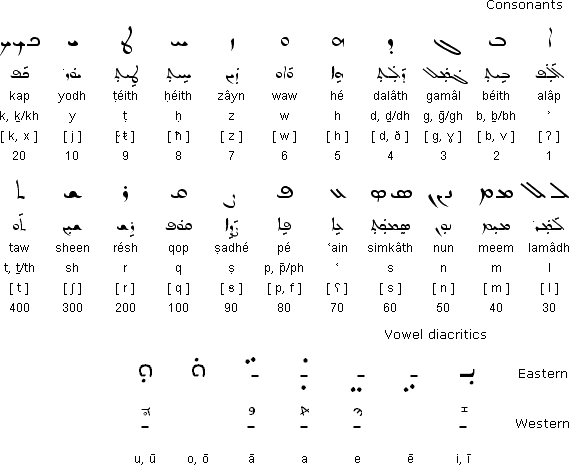}}
    \caption{West Syriac (Serṭā) Script \citep{omniglot}}
\label{fig3}

\end{figure}

As specified by UNESCO World Atlas of Languages \citeyearpar{unesco}, most Aramaic dialects that use the Syriac alphabet are classified as endangered; therefore, digital resources and tools are of extreme significance for the preservation and survival of the language and its cultural heritage in the technological era. Developing a handwritten OCR model for Syriac is a difficult task as it is a low-resourced and computationally unavailable language, yet it is highly needed for the purpose of digitizing vast amounts of handwritten manuscripts, many of which may date back to the first millennium; and it is also required as a stepping stone for providing more common digital services for the language and its users. Syriac is a low-resource and also endangered language due to centuries of geopolitical conflict and lack of policy to preserve the language by the governments of the aforementioned countries that hold the most population of Syriac speaking people. For instance, only until after 2005 did the Iraqi Federal Government \citep{iraqcon} recognize Syriac as one of the official languages in administrative units where the population is mostly christian communities; thus, a shortage in both digital services and academic research in the field of artificial intelligence can be witnessed since the language was not publicly used or taught in schools before 2005.
Syriac is a dialect of the Aramaic language from the greater family of Semitic languages which is said to have originated in or around Edessa \citep{butts} and was, and still is in its newer forms, mainly spoken by Christians of the Middle East, especially in Iraq, Iran, Turkey, and Syria. The Syriac alphabet consist of 22 letters and is written from right to left in a cursive style \citep{ackroyd}. Vowels are represented with diacritics called zāw'ā, and they are added above or below a character to alter the consonant's pronunciation; however, for the purpose of this paper, all form of diacritics will be omitted during data collection so as to make the process of character recognition easier and to avoid further ambiguity. There are three main writing systems in Syriac, namely Estrangela, East Syriac (Madnḥāyā), and West Syriac (Serṭā) \citep{omniglot}.  The letters kap and nun have three variations as shown in \autoref{fig}, \autoref{fig2}, and \autoref{fig3} where the one on right being used in the beginning of a word, middle one used at end of a word, and the one left used in final positions when unconnected to a previous letter. However, the letter kap has different shape only when in initial position in a word using the Estrangela script. Also, the letters meem and simkâth have one variation of writing for initial and middle position and one for final positions. Furthermore, the letter lamâdh differs in West Syriac in the beginning of a word when compared to East Syriac and Estrangela.  East Syriac will be the script used for this paper as it was most common amongst the Christian community in Iraq and it was underrepresented in most research papers. It shares more similarities with Estrangela than with West Syriac and it is also commonly referred to as Assyrian, Chaldean, or Nestorian.

\section{Related Works}\label{sec3}

This literature review sets to explore related works that primarily focus on handwritten OCR for cursive writing, and it will be divided into two sections: analytical approach and holistic approach. The fundamental aim of this Section is to reach a conclusion as to which method is most feasible and guarantees high rate of recognition based solely on the previous researches done in Syriac, Arabic, and Farsi.

\subsection{Analytical Approach}
One of the earliest attempts for Syriac handwritten OCR used both whole words and character shapes for recognition of Estrangela script, and the experiment results ranged "from 61-100\% depending on the algorithm used and the size and source of the dataset" \citep{clocksin}. A novel probabilistic method for segmentation was implemented to split each word into separate characters by scoring each pixel in a word with a likelihood of being a valid segmentation point, and it is recognized by the aid of a discriminative support vector machine (SVM) with 10-fold cross-validation. However, the segmentation approach failed in correctly splitting some letters, especially in two distinct scenarios: the first being the letter nun which crosses the baseline that was set by the segmentation algorithm, and the second letter being ḥeith because its misclassified as two separate letters due to its shape. Furthermore, another limitation this system has is that it is entirely "knowledge-free", i.e, no structural information of Syriac letter shapes or statistics have been used.

In contrast to the previous method, \cite{oversegment} proposed a segmentation algorithm for handwritten Arabic which "over-segments each word, then removes extra breakpoints using knowledge of the letter shapes". The segmentation method achieved a 92.3\% accuracy in detection of segment points on 200 test images of Tunisian city-names from the IFN/ENIT database, and over-segmentation occurred on 5.1\% of the instances. Furthermore, a neuro-heuristic algorithm was utilized for Arabic characters by \cite{hamid}. It initially segments the text into a block of characters using any conventional algorithm, which would then generate a presegementation for these blocks. In the end, a neural network verifies the accuracy of the segmentation points. This approach achieved an accuracy of 53.11\%, and 69.72\% after rejecting 35.43\% of the points.
Similarly, a deep learning recognition-based segmentation approach has been implemented for Syriac characters using a convolutional neural network (CNN) as a classifier alongside a variable-size sliding window that goes through each word image \citep{fermanian}. The CNN was trained on the DASH Project\footnote{https://dash.stanford.edu/} dataset using different classifiers and resulted in an accuracy between 94\% and 96\% depending on the CNN architecture used.

\subsection{Holistic Approach}
The problem of segmentation and the act of separating each entity into smaller pieces has troubled researchers who have been working on OCR, especially for cursive scripts. Sayre's paradox states that a handwritten cursive word cannot be segmented unless recognized, and it also cannot be recognized unless segmented \citep{sayre}; therefore, holistic methods could be taken into consideration as they provide a segmentation-free solution to handwritten OCR and provide recognition on the word-level. Nevertheless, it may reach a bottleneck since it relies on a large lexicon of word images, and it will recognize a word only insofar as the model has been exposed to it during training. \cite{bilane} were the first to bring forth a segmentation-free method for word spotting in West Syriac manuscripts that is based on a selective sliding window technique, from which it extracts directional features and then performs a matching using Euclidean distance correspondence between features. This approach required no knowledge of the language and its letter shapes, and despite the fact that no quantitative evaluation for the performance of this method has been presented and the documents lacked a ground-truth, it was still able to recognize frequently repeated words in the manuscripts. \cite{khorsheed} proposed a holistic approach for recognition of handwritten Arabic script. A single hidden Markov model (HMM) has been trained with the structural features extracted from the words of the manuscript. The HMM is composed of multiple character models where each model constitutes one letter from the Arabic alphabet. The recognition rates of the top-1 results were 72\% without spelling check and 87\% with spelling check, and the top-5 rates were 81\% without spelling-check and 97\% with spelling check.

Moreover, another research leveraging HMMs for holistic recognition of handwritten Farsi words was trained on a dataset of 17820 word images divided into 198 sections, each section representing a city in Iran \citep{farsihmm}. The proposed method's top choice accuracy was 65\% even though no contextual information was used to aid with the recognition.
In more recent literature, neural network approaches have been suggested for handwritten recognition. \cite{blstm} presented a hybrid approach for handwritten Arabic based on CNN to automatically extract features from raw image and a bidirectional long short-term memory (BLSTM) followed by a connectionist temporal classification (CTC) for sequence labelling. The data used for training this model was the IFN/ENIT database, which had data augmentation techniques applied on the samples to fit the use of the training model. According to the experiments done on the aforementioned dataset, this method resulted in an accuracy of 92.21\%. Also, \cite{dropout} claim that in order to avoid overfitting and overgeneralizing the training data, two dropout layers are added that drops 50\% of nodes each time, of which reduced label error rate by more than 4.88\%.
Additionally, based on the work of \cite{amrouch}, a tandem combination of both CNN and HMM has been suggested for Arabic handwritten recognition, where the CNN serves as a feature extraction layer and the HMM as a classifier. This model was also trained on the IFN/ENIT database of word images, and it resulted in an accuracy range of 88.95-89.23\%.

\subsection{Summary}
Comparison of both holistic approaches and analytical approaches can be tedious and time consuming as there is an abundance of literature in both, and they propose methods that achieve high performance in the two approaches. However, for the scope of this research, Tesseract-OCR engine will be utilized to develop the model. Details about the methodology and Tesseract-OCR will be discussed in the next Section.

\section{Methodology}\label{sec4}
This Section will discuss the methods and procedures that will be performed to develop an OCR model for handwritten Syriac text, and it will argue their significance for the success of the research's objective.

\subsection{Data collection}
Publicly available image datasets can be of massive importance for both research usage and industry applications, for they serve as a benchmark for the performance and strength of any image recognition model. Since the work is dealing with handwritten recognition and open-source dataset for handwritten Syriac are scarce, especially for East Syriac script, a handwritten Syriac dataset will be collected from scratch and it is partially published until future notice where the full dataset will be available. The dataset template form will consists of multiple sentences, of which each sentence has a bounding box beneath it so that the volunteer/writer will fill the specified area for writing. In addition, a pilot will be performed and each form will be given to volunteers who are capable of reading and writing in Syriac, specifically individuals in universities, of which then the form papers will be scanned using a digital scanner.

\subsection{Preprocessing}
After each form paper has been scanned, the raw images will go through a preproccesing phase so as to make the input data as intelligible and readable as possible for the classifier during model training. Initially, the bounding box of each sentence will be extracted from the form paper, including its content. One point to consider is that of noise, of which, according to \cite{lyall}, makes handwritten recognition more challenging due to the different writing tools, quality of the papers that has been used, and the pressure applied while the text was being written, which greatly dictate the quality of the input. A viable solution to this issue would be average blurring, which softens edges, noise, and other high frequency content in an image by convolving an image with a normalized box filter through taking the mean of all the pixels under the kernel area and replacing the central element through the use of OpenCV\footnote{https://opencv.org/}, which is an open source computer vision and machine learning software library. \autoref{eq1} shows a simple blurring mechanism of a 4x4 box filter \citep{opencv}.

\begin{equation}
    K = \frac{1}{16} \begin{bmatrix} 1 & 1 & 1 & 1\\ 1 & 1 & 1 & 1 \\ 1 & 1 & 1 & 1 \\ 1& 1 & 1 &1 \end{bmatrix}
    \label{eq1}
\end{equation}

Additionally, based on the premise that different writing tools will be used, images will be converted to grayscale and binarization may be required through using a thresholding algorithm which will quantize a pixel of an image to its maximum value based on a threshold and anything below that range will be dropped to zero as per shown in \autoref{eq2} \citep{gonzalez}.

\begin{equation}
g(x,y) = \left\{\begin{matrix}
1 & & f(x,y)>T\\ 
0 & & f(x,y) \leq T
\end{matrix}\right
.\label{eq2}
\end{equation}

\subsection{Tesseract-OCR}
According to Tesseract's documentation \citep{tesseract}, this OCR engine "was originally developed at Hewlett-Packard Laboratories Bristol UK and at Hewlett-Packard Co, Greeley Colorado USA between 1985 and 1994, with some more changes made in 1996 to port to Windows, and some C++izing in 1998. In 2005 Tesseract was open sourced by HP. From 2006 until November 2018 it was developed by Google." It is currently maintained by community contributors. Besides, Tesseract has "out of the box" support for more than a 100 languages, and it can be trained to recognize newer languages using the Tesseract tesstrain training tool that is based on a Long Short-Term Memory (LSTM) neural network architecture, which is a form of recurrent neural network that carries the weight of the previous layer to  preserve the sequential nature of the data. In the case of having small amount of samples in the dataset, Tesseract has support for fine-tuning pretrained models which could be utilized to train the target using weights and abstract information that the base model has already been exposed to to mitigate the issue of data scarcity.

\subsection{Train/Eval Split} After the raw image data has been preprocessed and prepared, and in order to avoid overfitting, the cleaned image dataset will be divided into two sets and undergo a 90/10 split, 90\% of the data being for training the model and 10\% designated for evaluation while training. Nevertheless, in the case of having small amount of data, a 80/20 and 70/30 split might be considered.

\subsection{Evaluation Criteria} As for evaluation, both character error rate (CER) and word error rate (WER) will be leveraged so as to compare the correctness of each recognized character or word to its ground truth value as per shown in the \autoref{eq3} and \autoref{eq4}.

\begin{equation}
CER = \frac{S + D + I }{N} \times 100 
.\label{eq3}
\end{equation}
\hspace{50pt}
\begin{equation}
WER = \frac{S + D + I }{N} \times 100
.\label{eq4}
\end{equation}

 $S$ indicates the number of character or word substitutions, $D$ referring to the deletions, $I$ being the incorrect character or word inserted during recognition, and $N$ being the total number of characters in the sample. It is also worth noting that if the model evaluation has a relatively low CER, it usually results in a slightly higher WER due to the fact that if a character in a recognized word is wrong it will assign the whole word as incorrect; thus, CER will be the main metric for the model assessment \citep{cer}.

\section{KHAMIS Dataset}\label{sec5}
\begin{figure}[H]
    
    \centerline{\includegraphics[width=80mm,scale=1]
    {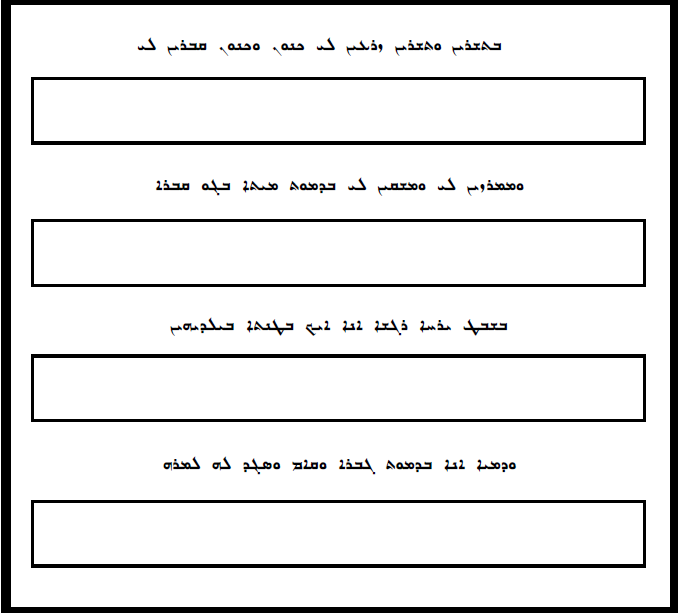}}
    \caption{Sample of a page from the dataset template form}
    \label{fig20}
\end{figure}

KHAMIS is the name of the dataset that was collected for the purpose of this research that encompass handwritten sentences based on a poem of Khamis bar Qardahe, who is a 13th century East Syriac poet and priest and is said to have been from Arbela (modern day Erbil, Iraq) \citep{khamis}. The dataset template form used to gather the samples consists of 20 sentences, of which a sentence is a verse from the poem as shown in \autoref{fig20}.

The data collection pilot was performed in between 31 university students and one professor who were capable in reading and writing in Syriac. Each volunteer was handed a dataset template form for input in designated bounding boxes, i.e, each volunteer was responsible of providing 20 handwritten sentences on that paper, however there were cases where less samples were provided. Moreover, the first batch of scanning, which were from all the participants, were scanned using an Epson WF-C5790 scanner; meanwhile, the second batch which was a re-scan of 6 participants' written A4 papers were scanned using a Xerox WorkCentre 7845 scanner. It is worth noting that the reason for the re-scanning of some raw samples were due to false configurations and low resolution settings in the first scanner, for which happened to not recognize or clearly scan image text that were not bold, especially those that were written using a pencil. 

\begin{figure}[H]
\label{fig50}
    \centerline{\includegraphics[width=80mm,scale=1]
    {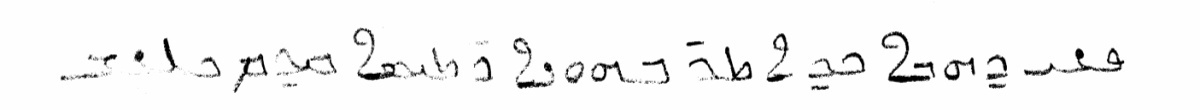}}
    \caption{Unclear sentence image}
    
\end{figure}

\begin{figure}[H]
\label{fig60}
    \centerline{\includegraphics[width=80mm,scale=1]
    {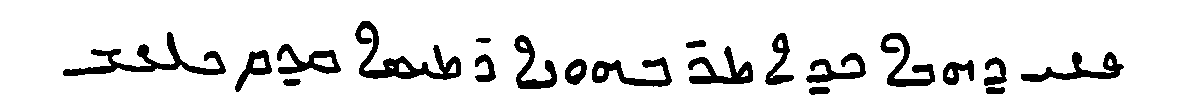}}
    \caption{Above sentence image after re-scan and preprocessing}
    
\end{figure}

Also, some sentence samples contained noise or words being scratched out or placed incorrectly; thus, manual sentence correction and noise removal was required either by repositioning the incorrectly placed segments or replacing unintelligible character or word with an identical one, or by removing the incorrect word or character from both the image and the ground-truth text file. 

\begin{figure}[H]
 \label{fig70}
    \centerline{\includegraphics[width=80mm,scale=1]
    {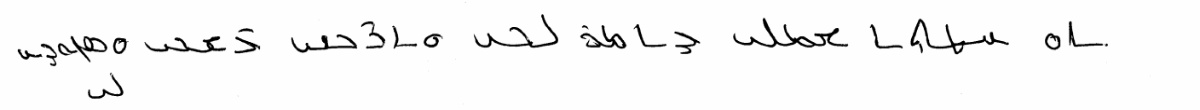}}
    \caption{Incorrectly placed sentence image}
   
\end{figure}

\begin{figure}[H]
\label{fig80}
    \centerline{\includegraphics[width=80mm,scale=1]
    {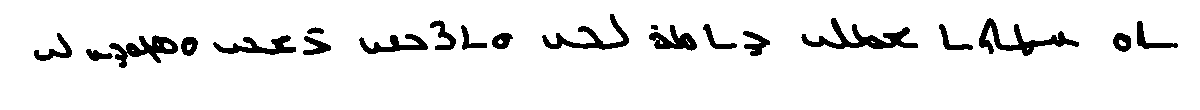}}
    \caption{Above sentence image after manual repositioning and preprocessing}
    
\end{figure}

In result, 624 1200x110 sentence image samples were collected, each accompanied with a text file containing its ground-truth value. Other than the data used to training the model, metadata about the volunteers have been collected with their permission, namely their age, gender, occupation, and their city or town of origin as shown in \autoref{meta}. It is worth mentioning that 13 of the participants were female and 19 were male with a mean age of 22.5 years old. Majority of participants were originally from Al-Hamdaniya/Bakhdida and Qaraqosh, however some individuals were from Duhok, Ainkawa, and Hakkari.

This is how the dataset folder is formatted: the image files being in PNG format with a name indicating the author and image number (those with letter "a" at the beginning are the first batch of scan and the ones with "b" are from the second batch, i.e., the re-scan), and a text file containing the ground-truth with a .gt.txt extension as suggested by Tesseract's Tesstrain documentation\footnote{https://github.com/tesseract-ocr/tesstrain}.

Tesseract initially uses Leptonica\footnote{https://tesseract-ocr.github.io/tessdoc/ImproveQuality.html} library for image preprocessing as a preceding step before the recognition process. OpenCV was used prior to Tesseract for extracting the raw sentence images from their bounding boxes in the original A4 scans and preprocessing them through binarization and noise removal using average blurring and a simple thresholding algorithm respectively, as stated in the previous Section.

\newpage

\begin{table}[t]
\caption{Volunteer metadata}

\begin{tabular}{@{}llll@{}}
\toprule
\textbf{Gender} & \textbf{Age} & \textbf{Educational Background} & \textbf{Place of Origin} \\ 
\midrule
Female & 23 & University Student   & Al-Hamdaniya/Bakhdida \\ 
Male   & 48 & University Professor & Ainkawa               \\ 
Male   & 18 & University Student   & Al-Hamdaniya/Bakhdida \\ 
Female & 18 & University Student   & Al-Hamdaniya/Qaraqosh \\ 
Female & 19 & University Student   & Duhok                 \\ 
Male   & 22 & University Student   & N/A                   \\ 
Female & 20 & University Student   & Al-Hamdaniya/Qaraqosh \\ 
Male   & 21 & University Student   & Hakkari               \\ 
Male   & 34 & University Student   & Al-Hamdaniya/Qaraqosh \\ 
Male   & 23 & University Student   & Duhok/Zakho           \\ 
Female & 21 & University Student   & Duhok                 \\ 
Female & 21 & University Student   & Ainkawa               \\ 
Male   & 18 & University Student   & Al-Hamdaniya/Bakhdida \\ 
Male   & 21 & University Student   & Ainkawa               \\ 
Male   & 21 & University Student   & Al-Hamdaniya/Bakhdida \\ 
Female & 21 & University Student   & Al-Hamdaniya/Qaraqosh \\ 
Male   & 20 & University Student   & Al-Hamdaniya/Qaraqosh \\ 
Male   & 23 & University Student   & Al-Hamdaniya/Bakhdida \\ 
Male   & 21 & University Student   & Al-Hamdaniya/Bakhdida \\ 
Male   & 19 & University Student   & Al-Hamdaniya/Qaraqosh \\ 
Male   & 25 & University Student   & Al-Hamdaniya/Bakhdida \\ 
Male   & 25 & University Student   & Al-Hamdaniya/Qaraqosh \\ 
Female & 21 & University Student   & Al-Hamdaniya/Bakhdida \\ 
Female & 23 & University Student   & N/A                   \\ 
Female & 22 & University Student   & Al-Hamdaniya/Bakhdida \\ 
Male   & 22 & University Student   & Al-Hamdaniya/Bakhdida \\ 
Female & 26 & University Student   & Al-Hamdaniya/Bakhdida \\ 
Female & 21 & University Student   & Al-Hamdaniya/Qaraqosh \\ 
Male   & 21 & University Student   & Ainkawa               \\ 
Female & 23 & University Student   & Al-Hamdaniya/Qaraqosh \\ 
Male   & 21 & University Student   & Al-Hamdaniya/Qaraqosh \\ 
Male   & 18 & University Student   & Al-Hamdaniya/Bakhdida \\ 
\bottomrule
\end{tabular}
\label{meta}

\end{table}

\section{Experiments and Results}\label{sec6}

    As mentioned in the previous Section, Tesseract leverages LSTM neural network architecture for training its models, and it also has the capability for fine-tuning, or to be more precise, using the weights of a pretrained model of a certain language with similar features to your target language. For this case, since Tesseract has already been trained on printed Syriac text using 20 fonts of the 3 different writing systems, mainly from the Beth Mardutho Institute's Meltho font package \footnote{https://bethmardutho.org/meltho/}, the learnt features could be utilized and fine-tuned in order to recognize handwritten samples from the desired data. Tesseract's Tesstrain training tool will be used for training on a Lenovo Thinkpad X1 Yoga, equipped with Linux Mint 21.3 Cinnamon operating system, Intel Core i7-8650U CPU, 512 GB SSD hard drive, and 16 GB RAM specifications.

    \newpage

\begin{table}[h]
    
           \caption{Device specifications for training model}
            \begin{tabular}{@{}ll@{}}
            \toprule
            \textbf{Computer Model} & Lenovo Thinkpad X1 Yoga  \\ 
\textbf{Operating System}      & Linux Mint 21.3 Cinnamon \\ 
\textbf{CPU}                   & Intel Core i7-8650U CPU  \\ 
\textbf{GPU}                   & Integrated Intel GPU     \\ 
\textbf{RAM}                   & 16 GB                    \\
\textbf{Hard Drive}            & 512 GB SSD               \\

            \bottomrule
        \end{tabular}

\end{table}

    The training process in Tesseract's Tesstrain is fairly straight forward insofar as you have the environment setup and dataset ready in the required format. The training configuration used for training the model are as showcased in \autoref{param}\footnote{The START\_MODEL is set to "syr", which indicates that the training will be starting from a checkpoint which is the pretrained model that has been trained on Syriac previously. The paramater "LANG\_TYPE" is set to "RTL" to inform the engine that the input used for training is a right-to-left script. For experimental purposes, three copies of the dataset has been created: esyr, esyr\textunderscore lesstrain, and esyr\textunderscore short, each being a "RATIO\_TRAIN" 0.9, 0.8, and 0.7 respectively.}.

    \begin{table}[h]
    
           \caption{Tesstrain training parameters}
            \begin{tabular}{@{}ll@{}}
            \toprule
            \textbf{Parameter} & \textbf{Value}\\
            \midrule
LEARNING\textunderscore RATE & 0.0001  \\ 
MAX\textunderscore ITERATIONS     & 10000 \\ 
START\textunderscore MODEL & syr  
\\ 
LANG\textunderscore TYPE   &  RTL    
\\ 
RATIO\textunderscore TRAIN &  (0.9, 0.8, 0.7)         \\

            \bottomrule
        \end{tabular}
       
\label{param}
\end{table}

\subsection{Training Results}
The performance of the proposed model were relatively high on both training and evaluation sets of all three variations and achieved a character error rate of 1.610\%, 1.402\%, and 1.097\% on training set, meanwhile it scored 9.864\%, 8.963\%, and 10.498\% on evaluation set, each value representing their respective groups in \autoref{tab}. 
\begin{table}[h]

                  \caption{Model training and evaluation set performance}

            \begin{tabular}{@{}llll@{}}
            \toprule
            \textbf{Name} & \textbf{Split} & \textbf{CER (Train)}  &  \textbf{CER (Eval)} \\ 
            \midrule
            esyr   & 90/10                     & 1.610\%                                  & 9.864\%                                    \\ 
esyr\_lesstrain & 80/20                     & 1.402\%                                  & 8.963\%                                    \\ 
esyr\_short     & 70/30                     & 1.097\%                                  & 10.498\%       \\
            \bottomrule
        \end{tabular}
        \label{tab}
\end{table}

\subsection{Evaluation on Test Data}
The performance of the model will be assessed on a test dataset that contains 12 sentence images and one paragraph image (13 samples in total). These samples have been collected by individuals who were not from the initial group of 32 volunteers, and they were also asked to slightly modify and add noise to the given sentences by replacing a word with their names or replacing multiple random characters with another character.

\begin{figure}[H]
 \label{fig30}
    \centerline{\includegraphics[width=80mm,scale=1]
    {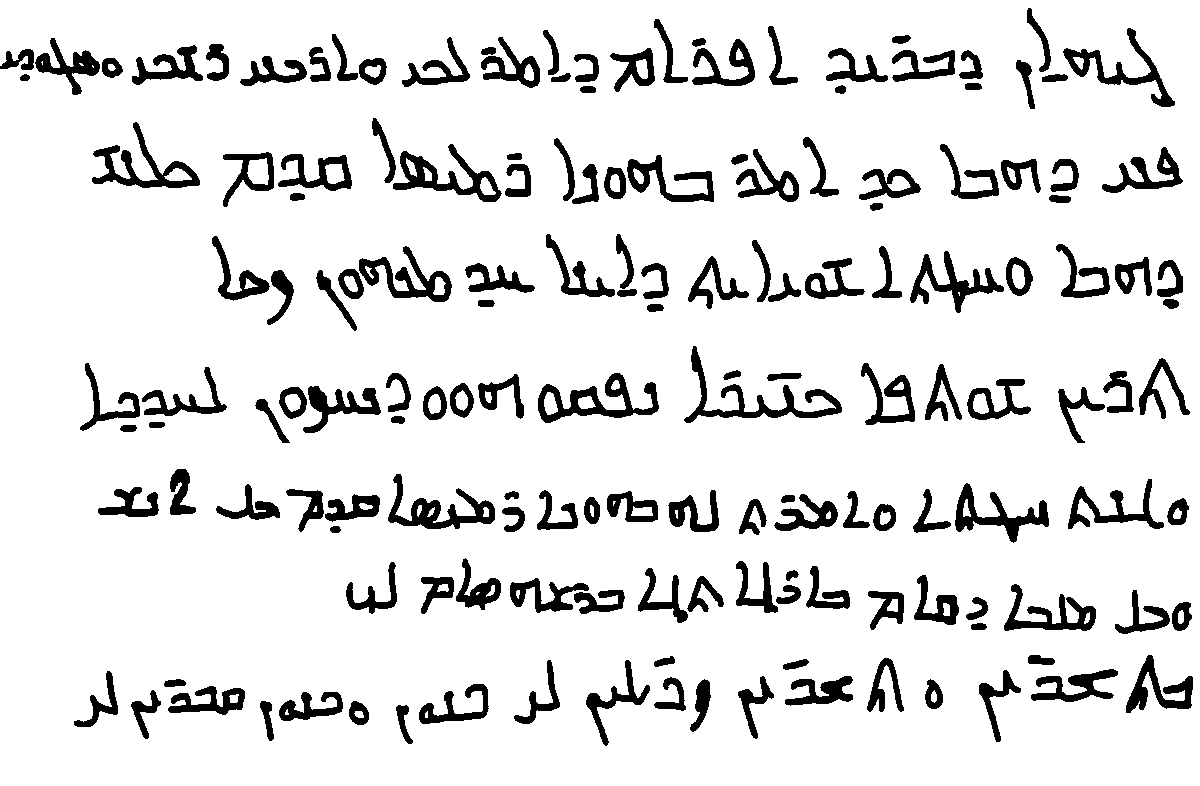}}
    \caption{Paragraph from test dataset}
   
    \end{figure}

\begin{figure}[H]
  \label{fig33}
    \centerline{\includegraphics[width=80mm,scale=1]
    {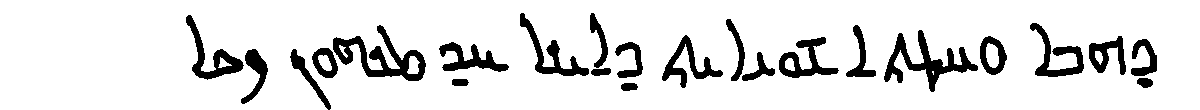}}
    \caption{Sentence from test dataset}
  
    \end{figure}

The mean of the character error rate (CER) and word error rate (WER) of all samples have been calculated so as to have a more generalized view on the model's performance on unseen data. When doing a comparison between the default model of Syriac OCR by Tesseract and the model that was fine-tuned on the KHAMIS dataset with the three different configurations, there is an evident improvement when it comes to recognition, notably twice less prone to incorrect classification than the default model. To demonstrate what the model does, inference is being done to a test image sample on only the "esyr" configuration to compare the predicted result with the ground-truth value.

\begin{figure}[H]
    \centerline{\includegraphics[width=80mm,scale=1]
    {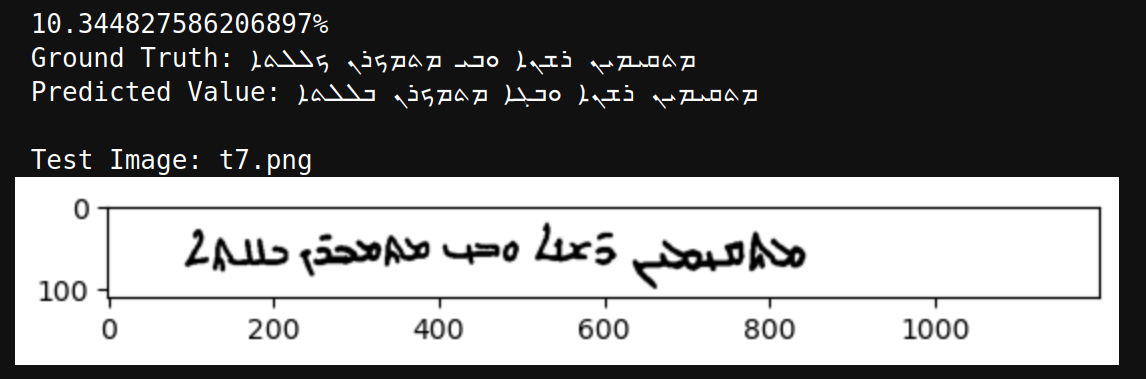}}
    \caption{Test image "t7.png" inference}
    \label{fig133}

    \end{figure}

\begin{figure}[H]

    \centerline{\includegraphics[width=80mm,scale=1]
    {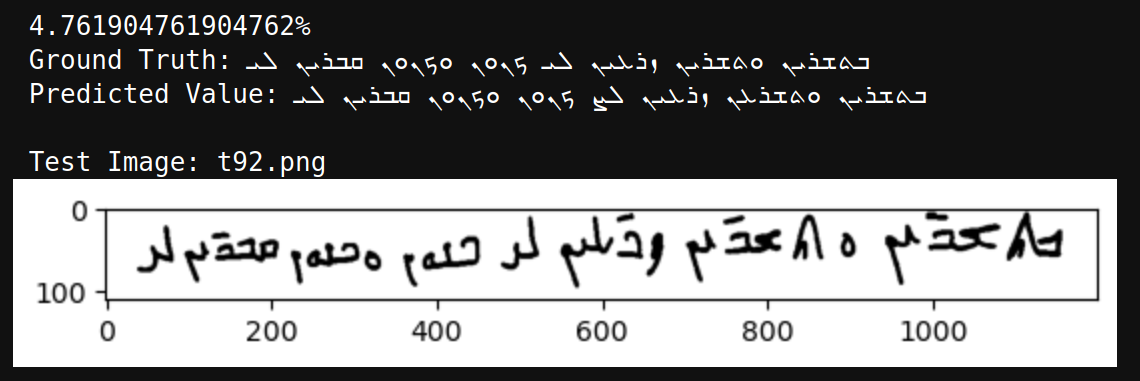}}
    \caption{Test image "t94.png" inference}
    \label{fig69}
    \end{figure}

The character errors that occurred in \autoref{fig133} are primarily within characters or a combination of two characters that looks similar to one. In this case it recognized the letter kap in the beginning of the final word as bēith, and yodh as the combination gāmal and ālap as shown in the third word. Additionally, in \autoref{fig69} the error that occured was the letter yodh being mistaken for ṣade in the fourth word, and another instance of yodh being mistaken for gāmal in the second word.

\begin{table}[h]

            \begin{tabular}{@{}lll@{}}
            \toprule
            \textbf{Name} & \textbf{CER (Test)}  &  \textbf{WER (Test)} \\ 
            \midrule
            Default Model: syr & 55.71\% & 122.78\%\\
            esyr & \textcolor{red}{19.80\%} & \textcolor{red}{64.41\%} \\
            esyr\_lesstrain &  \textcolor{red}{18.82\%} & \textcolor{red}{62.83\%} \\
            esyr\_short & \textcolor{red}{19.71\%}  & \textcolor{red}{65.42\%} \\
            \bottomrule
        \end{tabular}
        \caption{Model test set performance}

\end{table}

\section{Discussion}\label{sec7}
Although this model was, nonetheless, able to showcase a relatively decent performance considering it was fine-tuned on a pretrained model using limited amount of data from what was collected for the KHAMIS dataset, creating a highly accurate OCR model, especially for an endangered and a computationally unavailable language like Syriac, requires vast amounts of image samples. One of the main limitations of this model was that it was trained on only 624 sentences; therefore, data augmentation techniques can be leveraged to both extend the length of the dataset and create different variations and positioning of the image samples. Besides, the evaluation of the model was done on a rather small test set, from which it could not give a concise representation of the accuracy of the model on data that it has not been exposed to. Moreover, such model could have presented more a respectable performance when trained on less computationally intensive machine learning algorithms rather than artificial neural networks (ANN), be it Hidden Markov Models (HMM), Support Vector Machines (SVM), etc.. Furthermore, this model will experience a bottleneck during inference of different scripts like Estrangela and West Syriac since it was only trained on East Syriac script; therefore, a more universal Syriac OCR model is recommended such that it is able to differentiate between the various scripts and their differences on the character level, and output a highly accurate result in all scenarios. Lastly, the inclusion of diacritics are of extreme importance as most contemporary documents or historical manuscripts do include them as many words change meaning when accompanied with different diacritics. 

\raggedbottom

\section{Conclusion and Future Works}\label{sec8}
Syriac is a language which is both historically and linguistically valuable, and its
survival is dependent on the people who participate in discourse and contribute to
society through the means of that language. This research’s focus was to develop
a handwritten OCR model for Syriac, and also to contribute to further research and
development projects and to mitigate the issue of data unavailability through creating
a handwritten dataset named KHAMIS. The proposed OCR model was able to achieve a character error rate of 1.097-1.610\% and 8.963-10.490\% on both training and evaluation sets respectively. Moreover, once evaluated with the test data samples they achieve both a character error rate of 18.89-19.71\% and a word error rate of 62.83-65.42\%, which is twice as better than the default Syriac model of Tesseract.

As a final remark, recommendation for a better handwritten OCR model in the future may be through such methods:

\begin{itemize}
    \item  More data collection through crowdsourcing projects or national \& international digitization initiatives within the Chaldean/Assyrian/Syriac community
 \item Recognition of the two other writing systems: (Estrangela and West Syriac)
 \item Diacritics to be included in future data samples
 \item Experiment with different algorithms and training parameters
\end{itemize}

\bmhead{Acknowledgements}
We are incredibly grateful to the Department of Syriac Education at Salahaddin University - Erbil and all their students who volunteered in the data collection process.












\bibliography{sn-bibliography}

\end{document}